%% file: root.tex
\documentclass[10pt,twocolumn,a4paper]{article}
\usepackage{cvpr} 
\include{src/headers}

\title{\LARGE \bf 
AquaFuse: Waterbody Fusion for Physics Guided\\ View Synthesis of Underwater Scenes 
\vspace{-2mm}
}

\author{Md Abu Bakr Siddique$^1$, Jiayi Wu$^2$, Ioannis Rekleitis$^3$, and Md Jahidul Islam$^4$% <-this % stops 
\\ 
Project page: \url{https://robopi.ece.ufl.edu/aquafuse.html}
\thanks{This work is accepted for publication at IEEE RA-L (03.2025)}
\thanks{$^{1,4}$RoboPI laboratory, Dept. of ECE, University of Florida, US}% 
\thanks{$^{2}$Dept. of ME, University of Delaware, Newark, DE, US}
\thanks{$^{3}$Dept. of CSE, University of South Carolina, Columbia, SC, US}
\vspace{-2mm}
}

\begin{document}

\maketitle
%\thispagestyle{empty}
%\pagestyle{empty}

%%%%%%%%%%%%%%%%%%%%%%%%%%%%%%%%

\input{src/Abstract}

\input{src/Introduction.tex}

\input{src/RelatedWork.tex}

\input{src/Methodology.tex}

\input{src/Experiments.tex}

\input{src/Conclusion}

{\small
\bibliographystyle{abbrv}
\bibliography{root}
}

\end{document}

%% file: src/headers.tex
\usepackage{cite}
\usepackage{amsmath,amssymb,amsfonts}
\usepackage{graphicx}
\usepackage{multirow}
\usepackage{physunits}

\DeclareMathOperator*{\argmin}{arg\,min}
\usepackage[colorlinks,bookmarksopen,bookmarksnumbered,citecolor=green,urlcolor=red]{hyperref}
\usepackage{makecell}
\usepackage{enumitem}
\usepackage{subcaption}
\usepackage[size=small]{caption}
\usepackage{booktabs}

%% file: src/Abstract.tex
\begin{abstract}
\vspace{-2mm}
We introduce the idea of \textbf{AquaFuse}, a physics-based method for synthesizing \textit{waterbody properties} in underwater imagery. We formulate a closed-form solution for waterbody fusion that facilitates realistic data augmentation and geometrically consistent underwater scene rendering. AquaFuse leverages the physical characteristics of light propagation underwater to synthesize the waterbody from one scene to the object contents of another. Unlike data-driven style transfer, AquaFuse preserves the depth consistency and object geometry in an input scene. We validate this unique feature by comprehensive experiments over diverse underwater scenes. We find that the \textit{AquaFused images}  preserve over 94\% depth consistency and 90-95\% structural similarity of the input scenes. We also demonstrate that it generates accurate 3D view synthesis by preserving object geometry while adapting to the inherent \textbf{waterbody fusion} process. AquaFuse opens up a new research direction in data augmentation by geometry-preserving style transfer for underwater imaging and robot vision applications.  \\

\noindent
\textit{\textbf{keywords}}--View Synthesis; Robot Vision; Image Processing.
\end{abstract}

\vspace{-5mm}

%% file: src/Introduction.tex
\vspace{-2mm}
\section{Introduction}
Light attenuates exponentially underwater with propagation distance~\cite{akkaynak2018revised,yu2022udepth} due to two physical characteristics of light: \textit{scattering} and \textit{absorption}. Forward scattering is responsible for blur, whereas backscatter causes contrast reduction~\cite{yang2019depth,islam2020fast}. Besides, light absorption by water has a prominent spectral dependency, which depends on the specific optical properties of a waterbody~\cite{akkaynak2019sea}, distance of light sources, salinity, and many other factors. These physical characteristics are represented by the underwater image formation model~\cite{akkaynak2018revised}, which, in recent years, has been successfully applied for water removal, image restoration~\cite{akkaynak2019sea}, and various scene rendering tasks~\cite{levy2023seathru,yang2024seasplat}. In particular, underwater scene synthesis approaches extend the state-of-the-art (SOTA) neural rendering methods such as  NeRF~\cite{mildenhall2021nerf} and Gaussian splatting~\cite{kerbl20233d} for domain-aware learning. They incorporate the physical properties of underwater image formation into the rendering pipeline for accurate view synthesis~\cite{levy2023seathru,yang2024seasplat}. This helps compensate for the inherent attenuation and optical artifacts in underwater imagery and ensures geometrically consistency. Various image recognition models and 3D reconstruction pipelines also use them for improved visual perception~\cite{yang2019depth,wu2023sdu_sfm}.

\begin{figure}[t]
    \centering
    \includegraphics [width=\linewidth]{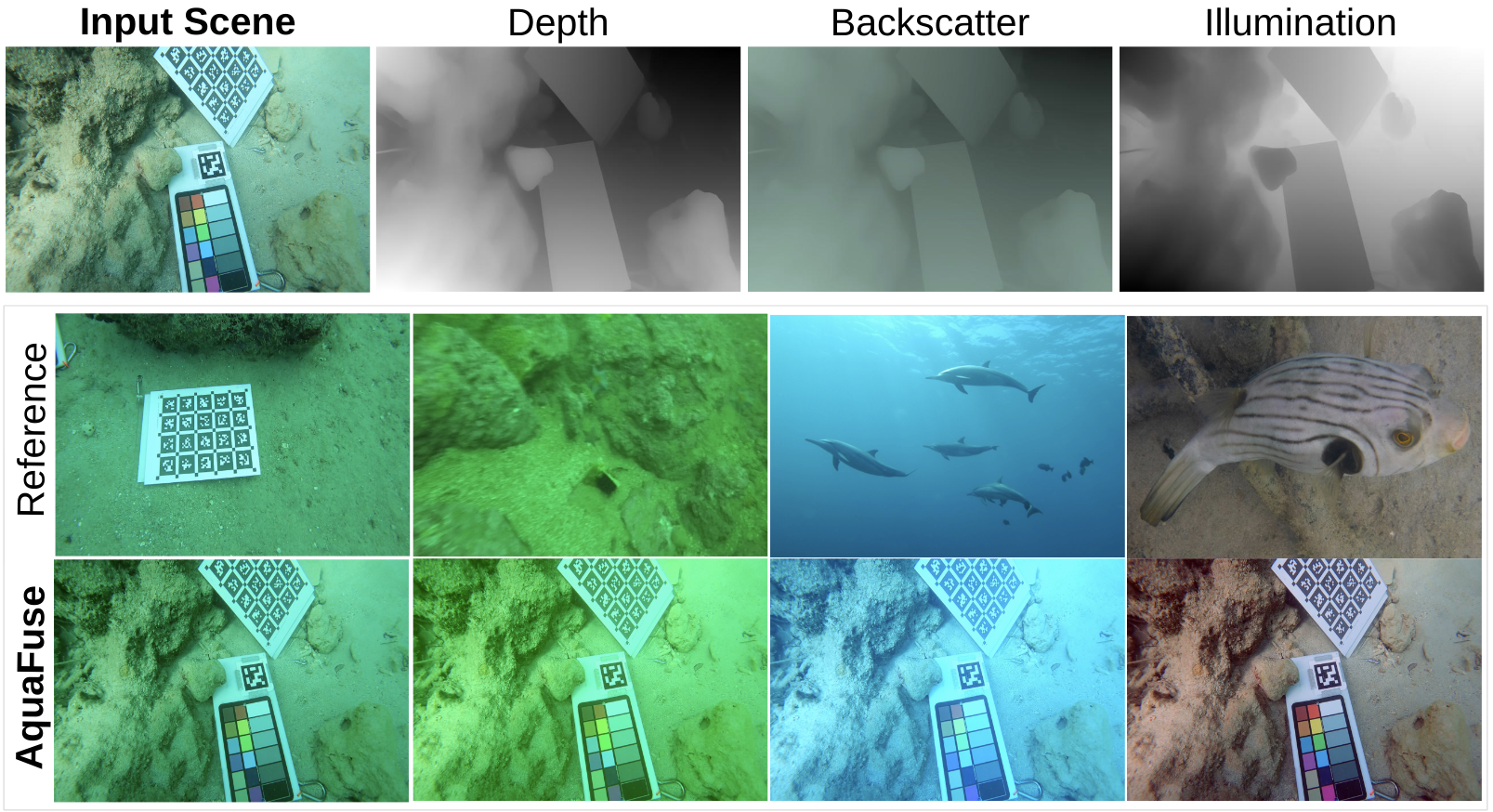}%
    \vspace{-3mm}
    \caption{AquaFuse is a physics-guided waterbody fusion method to \textit{fuse} waterbody properties of a reference image to an input image. It leverages closed-form solutions to estimate and exploit the scene depth, background light, backscatter, and medium illumination parameters for guiding the fusion process. As shown, AquaFused scenes are perceptually realistic and also meet the underwater image formation constraints.}%
    \vspace{-2mm}
\label{fig:intro}
\end{figure}

In this paper, we explore a new technique for domain-aware data augmentation by fusing waterbody across different underwater scenes. The traditional approaches of data augmentation rely on preserving perspective or isometric constraints (\eg, rotations, translations, scaling, shear, dilation) or altering basic photometric properties (\eg, brightness, color, contrast, saturation). We hypothesize that physically accurate waterbody fusion can be an effective way to augment data for underwater image recognition and scene rendering research; see Fig.~\ref{fig:intro}.

Note that the image style transfer literature~\cite{chen2024upst,zheng2024fast} offers solutions that learn artistic or photometric features from data. 
These methods rely on generative or adversarial networks to adapt one image domain's style to another~\cite{fabbri2018enhancing,li2018watergan}. Existing learning pipelines focus on perceptual photorealism~\cite{chiu2022photowct2} or artistic blending~\cite{chen2021artistic,deng2021arbitrary,deng2022stytr2}), and thus cannot guarantee geometric preservation and depth integrity. While these models are computationally heavy themselves, running physics-based water removal methods~\cite{akkaynak2019sea} in the backend would make it infeasible for end-to-end and/or real-time use~\cite{islam2020fast}. Hence, data augmentation by such \textit{domain transfer} is not suitable if physical realism and object geometry preservation are crucial, as in 3D view synthesis, scene reconstructions, and robotic mapping tasks~\cite{yang2024seasplat,wu2023sdu_sfm}.

We address these issues in \textbf{AquaFuse} by leveraging the revised underwater image formation model (UIFM)~\cite{akkaynak2018revised,akkaynak2019sea} to \textit{fuse} the optical characteristics of various waterbody types while preserving the object geometry on the scene. We formulate a nonlinear domain projection method to estimate scene depth in real-time, which facilitates an empirical estimation of the veiling light and illumination map. Veiling light is the global {background light} content of the backscatter signal, which decays exponentially with depth (optical distance)~\cite{akkaynak2019sea,li2016single,li2018emerging}. We decouple this global component from a reference image and integrate it into another (input) scene for backscatter fusion. The fused background light and backscatter signal drive the waterbody fusion, while the direct attenuation and illumination map remain preserved. We then reconstruct the AquaFused image by following the revised UIFM, retaining its physical properties.

In addition to the theoretical analysis, we present the engineering constructs to tackle the practicalities of waterbody fusion across multiple scenes. We demonstrate how AquaFuse can be used for waterbody crossover, image enhancement by fusing with a clear reference scene, and in a generative pipeline for data augmentation. For qualitative and quantitative validation, we collect data at multiple ocean sites over diverse waterbody types and depth levels ($10'$-$80'$). Our analyses of $N$ images across six water body types show that \textit{AquaFused images}  preserve over $94\%$ depth consistency and $90$-$95\%$ structural similarity of input scenes. The performance margins vary $1$-$5$\% with increasing water depths from $10'$ to $80'$. We compare the performance with SOTA methods for traditional style transfer, including ASTMAN~\cite{deng2020arbitrary}, IEContraAST~\cite{chen2021artistic}, MCCNet~\cite{deng2021arbitrary}, PhotoWCT2~\cite{chiu2022photowct2} and StyTr2~\cite{deng2022stytr2}. With qualitative and quantitative analyses, we demonstrate where traditional learning pipelines fall short for waterbody fusion.

Moreover, we integrate AquaFuse within a 3D view synthesis pipeline using Gaussian Splatting~\cite{kerbl20233d} for underwater scene reconstruction. We find that AquaFuse can generate highly detailed 3D reconstructions that represent the original scene's geometric characteristics in the fused scenes. Experimental results show that AquaFuse (\textbf{i}) performs realistic and accurate waterbody fusion, validated by PSNR (peak signal-to-noise ratio), SSIM (structural similarity index measure), and LPIPS (learned perceptual image patch similarity) scores; (\textbf{ii}) ensures geometric consistency across underwater scenes, validated by SND (surface normal deviation); and (\textbf{iii}) is invariant to dataset bias and overfitting unlike learning-based approaches.

As annotated underwater image databases are scarce, physics-guided data augmentation by AquaFuse can facilitate a multi-fold increase in training samples, which can be used for both 2D image recognition and 3D scene reconstruction~\cite{yang2024seasplat,zhang2024recgs,levy2023seathru}. This covers a wide spectrum of robot vision tasks, empowering data-driven models to be more robust and generalizable~\cite{yu2022udepth,abdullah2023caveseg}. Moreover, being a closed-form solution, AquaFuse offers over $342$\,FPS ($20$\,FPS) inference for $256\times256$ ($1920\times1080$) resolutions on a Core\texttrademark{}i7 CPU. In fact, it runs on a Raspberry Pi-4 device at over $22$\, FPS ($256\times256$ resolution), demonstrating its utility for real-time active vision on embedded platforms, which no existing style transfer methods can offer.

%% file: src/RelatedWork.tex
\section{Background \& Preliminaries}\label{sec:background}
Underwater images are formed by two components: a direct signal containing the attenuated scene information ($\mathbf{D}$) and an additive backscatter signal ($\mathbf{B}$) caused by rebounded light reflected from suspended water particles. The widely used {Jaffe-McGlamery imaging model}~\cite{jaffe1990computer,mcglamery1980computer} is based on the {atmospheric scattering model}~\cite{he2010single,xue2021joint}; it is defined as:%%
\vspace{-1mm}
\begin{align*}
    \mathbf U_\mathbf \lambda(\mathbf x) &= \mathbf{D}_\mathbf \lambda(\mathbf x) + \mathbf{B}_\mathbf \lambda(\mathbf x) \\
    &= {\mathbf I_\mathbf \lambda(\mathbf x) \cdot e^{-\beta_\mathbf \lambda d(\mathbf x)}} + {\mathbf B_\mathbf \lambda^\infty \cdot \big( 1- e^{-\beta_\lambda d(\mathbf x)} \big). }\label{eq6}
\end{align*}%  
Here, $\mathbf U$ is the observed underwater image and $\mathbf I$ is the clear latent image we want to recover; $\mathbf \lambda$ denotes the different wavelengths for RGB channels and $d(\mathbf x)$ is the actual distance of point $\mathbf x$ from camera along the line-of-sight, approximated by depth estimators ($\mathbf z$). The illumination map of the medium is an exponential of the form ${\mathbf E_\mathbf \lambda} \sim t_\mathbf \lambda(\mathbf x) = e^{-\beta_\lambda \cdot d(\mathbf x)}$, with $\beta_\mathbf \lambda$ as the uniform attenuation coefficient. Lastly, $\mathbf B_\mathbf \lambda^\infty$ represents the background (veiling) light.

The {revised underwater image formation model}~\cite{akkaynak2018revised,akkaynak2019sea} shows that the effects of light absorption can be comparable to light scattering or even dominate it depending on the waterbody. The attenuation coefficient relies not only on the optical properties of water, but also on the sensor, scene irradiance, and imaging range. More importantly, attenuation coefficients of backscattering and direct signal are different:
$\beta_\mathbf \lambda^B = -\ln(1-{\mathbf B_\mathbf \lambda}/{\mathbf B_\mathbf \lambda^\infty})/\mathbf z$ 
and
$\beta_\mathbf \lambda^D = -\ln({(\mathbf U_\mathbf \lambda - \mathbf B_\mathbf \lambda)}/{\mathbf I_\mathbf \lambda})/\mathbf z$, respectively. The revised model is then expressed by: %
\vspace{-1mm}
\begin{equation}
    \mathbf U_\lambda(\mathbf x) = {\mathbf I_\lambda(\mathbf x) \cdot e^{-\beta_\mathbf \lambda^D (\mathbf v_D) \cdot \mathbf z}} + {\mathbf B_\mathbf \lambda^\infty \cdot (1 - e^{-\beta_\mathbf \lambda^B (\mathbf v_B) \cdot \mathbf z})} \label{eq9}
\end{equation}
where $\mathbf v_D = \{z, \rho, {E}, S_\lambda, \beta \}$ is the coefficient dependency of direct signal and $\mathbf v_B = \{{E}, S_\lambda, b, \beta \}$ is coefficient dependency of the backscattered signal. Here, $\rho$ is the reflectance, ${E}$ is the spectrum of ambient light, $S_\lambda$ is spectral response, $b$ is beam scattering coefficient, and $\beta$ is beam attenuation coefficient.

In recent years, this revised image formation model has been successfully used for rendering underwater scenes by methods such as SeaThru-NeRF~\cite{levy2023seathru}, WaterSplatting~\cite{li2024watersplatting}, and SeaSplat~\cite{yang2024seasplat}. These methods leverage the remarkable advances in 3D radiance fields, specifically NeRF~\cite{mildenhall2021nerf} and Gaussian splatting~\cite{kerbl20233d}-based rendering. However, drawbacks remain in accounting for multiple scattering or inconsistent/artificial illumination. They also require camera poses to be extracted beforehand, which is often not feasible in practice. Thus, collecting and synthesizing comprehensive data for 3D underwater scene reconstruction research is rather challenging.

%% file: src/Methodology.tex
\section{AquaFuse Methodology}\label{sec:pipeline}
The revised model in Eq.~\ref{eq9} accurately describes the underlying physical effects of absorption and scattering on underwater image formulation~\cite{akkaynak2018revised}. However, it requires dense scene depth, which is typically estimated by sub-optimal linear estimators or noisy predictors such as MonoDepth2~\cite{godard2019digging} and AdaBins~\cite{bhat2021adabins}. Moreover, due to strong depth-dependency of the attentuation coefficients, camera parameters such as the reflectance, irradiance, and spectral response are ignored in practice. With these simplifications, the model reduces to: 
\vspace{-1mm}
\begin{align*}%
\vspace{-2mm}
    \mathbf U_\lambda(\mathbf x) &= {\mathbf I_\lambda(\mathbf x) \cdot e^{-\beta_\mathbf \lambda^D (\mathbf z) \cdot \mathbf z}} + { { \mathbf B_\mathbf \lambda^\infty \cdot (1 - e^{-\beta_\mathbf \lambda^B (\mathbf z) \cdot \mathbf z})}}  \\
    &= {\mathbf I_\lambda(\mathbf x) \cdot e^{-\beta_\mathbf \lambda^D (\mathbf z) \cdot \mathbf z}} +  \mathbf B_\lambda(\mathbf x).
\end{align*}
Hence, the clear latent image can be reconstructed as: 
\begin{equation}%
\vspace{-2mm}
    \mathbf I_\lambda(\mathbf x) = [ {\mathbf U_\lambda(\mathbf x) - \mathbf B_\lambda(\mathbf x)] \cdot e^{\beta_\mathbf \lambda^D (\mathbf z) \cdot \mathbf z}} \label{reconstruct}.
\end{equation}

% \begin{figure*}[t]
%     \centering
%     \vspace{-1mm}
% \includegraphics[width=0.9\linewidth]{images/UIE_Pipeline.jpeg}%
%     \vspace{-2mm}
%     \caption{The end-to-end process of image recovery based on the revised UIFM's simplified form is shown. The depth map $\mathbf{z}$ is first estimated from the raw input image, and then further exploited to obtain the backscatter signal $\mathbf{B}$ and attenuation coefficient ${\beta^D}$. Finally, the recovered image $\mathbf{I}$ is reconstructed by substituting these parameters in the Equation~\ref{reconstruct}.}
%     \label{fig:thrust_1}
% \end{figure*}

\begin{figure}[h]
    \centering
    \includegraphics [width=\linewidth]{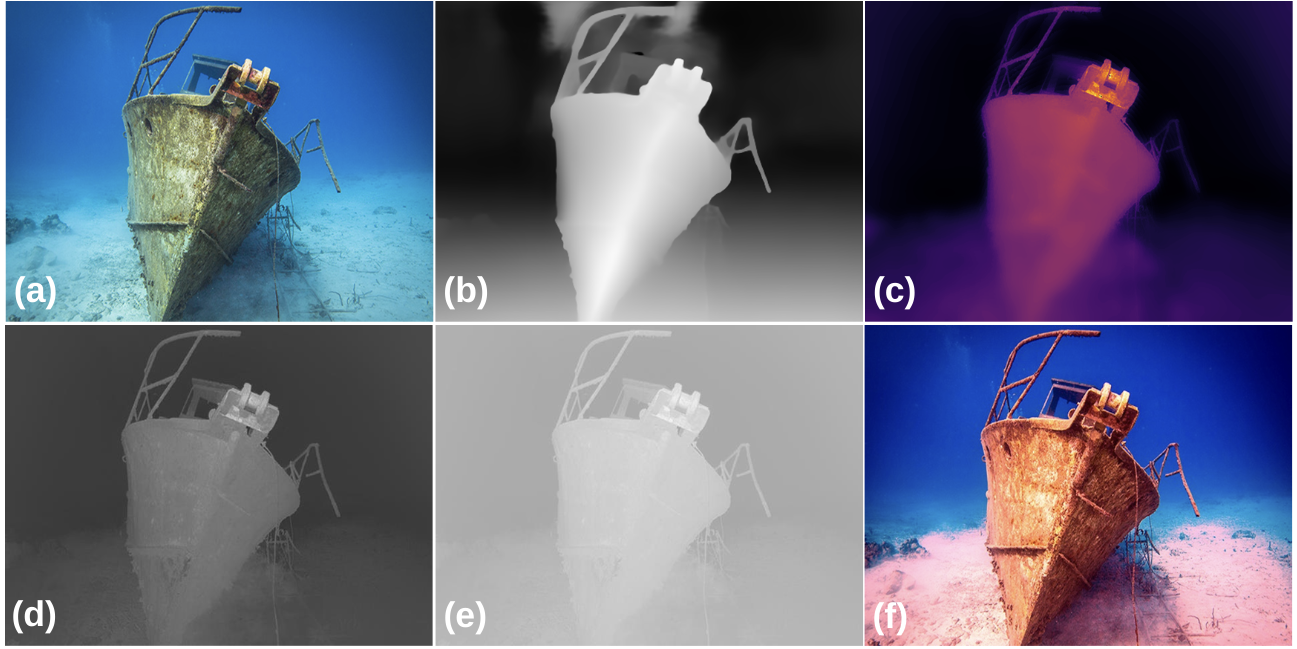}%
    \vspace{-2mm}
    \caption{An illustrative example of our image restoration pipeline; (a) input signal $\mathbf U$; (b) ground truth depth; (c) estimated depth $\mathbf Z$; (d,e) estimated illumination map for $R$ and $M$ channels; and (f) the restored signal $\mathbf I$.}%
    \vspace{-2mm}
\label{fig:WP_box}
\end{figure}

\subsection{Real-time Depth Estimation}
Since red wavelength suffers more aggressive attenuation underwater~\cite{galdran2015automatic,song2018rapid}, the relative differences between \{R\} and \{G, B\} values encode useful pixel-wise depth information for an image. Our prior work~\cite{yu2022udepth} exploits this to demonstrate that R\textbf{M}I$\equiv$\{R, \textbf{M}$=$$max$\{G,B\}, I (intensity)\} is a better input space for underwater scene depth estimation. Considering the exponential dependencies of the $R$ and $M$, we extend this idea for improved depth estimation with nonlinear estimators. We formulate the following two-term exponential for real-time depth estimation.
\begin{equation}
    \tilde{d}(\mathbf x \big| R, M) = \mu_0 + \mu_1 e^{\mu_2 \cdot R(\mathbf x)} + \mu_3 e^{\mu_4 \cdot M(\mathbf x)} + \epsilon(\cdot).
    \label{nle_depth}
\end{equation}
Here, $\mu$=$\mu_{0:4}$ are unknowns and $\epsilon(\cdot)$ is the residual error function that depends on the waterbody parameters. We use over $10$\,K natural underwater images of the USOD10K dataset~\cite{USOD10K} optimize $\mu$ by
\begin{equation}
 \mu^* = \argmin_\mu \sum_{i,\mathbf x} {\big\lVert {d}_{i}(\mathbf x) - \tilde{d}(\mathbf x \big| R_i, M_i) \big\rVert}_2^2.%
\label{rmi_opt}
\end{equation}%
A sample example of estimated $\mathbf Z \sim \tilde{d}(\cdot)$ is shown in Fig.~\ref{fig:WP_box}. In our implementation, an additional \textit{median blur} is applied to reduce noise and preserve edges, which ensures the robustness of background light estimation (see Sec.~\ref{subsec:3B}). Specifically, we perform
$\mathbf{z} = MedianBlur(\mathbf{z}, 7) \in \mathbf Z$ to filter the estimated depth map with a kernel size of $7$.

\subsection{Waterbody Parameter Estimation}
\label{subsec:3B}
An initial estimate of the backscatter signal is typically computed from the pixels where light is {totally absorbed}, \ie, by exploiting the lowest $1\%$-$5\%$ image intensities~\cite{akkaynak2019sea}. These image values correspond to black or completely shadowed regions ($\mathbf E$$\rightarrow$$0$). Besides, if $\mathbf z$ is large or for extremely turbid water, $\mathbf{B}_\mathbf \lambda$ increases exponentially and dominates $\mathbf{D}_\mathbf \lambda$. These cases are used to get an initial estimate $\mathbf U_\mathbf \lambda$$\rightarrow$$\mathbf{B}_\mathbf \lambda$. Moreover, we formulate the standard equation to estimate $\mathbf{B}_\mathbf \lambda$ from veiling light $\mathbf B_\mathbf \lambda^\infty$ is:
%For subsequent refinements, a non-linear least squares estimation of $\mathbf{B}$ can be modeled as:
\vspace{-1mm}
\begin{equation}%
    \tilde{\mathbf B}_\mathbf \lambda = \mathbf B_\mathbf \lambda^\infty (1 - e^{-\beta_\mathbf \lambda^B \cdot \mathbf z}) + \epsilon(\cdot)
    \label{NLE_B}.
\end{equation}
Here, $\mathbf B_\mathbf \lambda^\infty$$\in$[$0$, $1$] and residual $\epsilon(\cdot) = \tilde{\mathbf I}_\lambda \cdot e^{-{\tilde \beta_\lambda^D \cdot \mathbf z }}$. We model the dependency of ${\tilde \beta_\mathbf \lambda^D}$ on $\mathbf z$ by non-linear functionals $\phi_1$ and $\phi_2$ as: $\Tilde{\beta}_\mathbf \lambda^D (\mathbf z) = \mu_5 e^{\mu_6 \phi_1 (\mathbf z)} + \mu_7 e^{\mu_8 \phi_2(\mathbf z)}$, where the unknowns are computed by a standard line fitting algorithm.

Once backcatter signal is removed from input image, the illumination map (${\mathbf E}$) is approximated by using the depth-color consistency assumption~\cite{henke2013removing,ebner2013depth}. According to the revised model~\cite{akkaynak2018revised,akkaynak2019sea}, the depth-scaled illumination map provides a good initial estimate for ${\beta}_\mathbf \lambda^D$; \ie, ${\beta}_\mathbf \lambda^D (\mathbf z) = - \log{\Tilde{\mathbf E}_\mathbf \lambda (\mathbf z)} / \mathbf z$. 
We further refine this by optimizing the following function:
\vspace{-1mm}
\begin{equation}%
    \min_{\beta_\mathbf \lambda^D (\mathbf z) \ge 0} \mathcal{L}_z = \min_{\beta_\mathbf \lambda^D (\mathbf z) \ge 0} \big\lVert \mathbf z - (- \log{\Tilde{\mathbf E}_\mathbf \lambda (\mathbf z)} / \beta_\mathbf \lambda^D (\mathbf z))  \big\lVert_2^2
    \label{refine_beta}.
\end{equation}
Specifically, we formulate a joint iterative optimization of ${\mathbf E}$ and $\beta_\mathbf \lambda^D (\mathbf z)$. With the initial conditions: 
\begin{align}
    \beta_0^D (\mathbf z) &= \mu_5 e^{\mu_6 \phi_1 (\mathbf z)} + \mu_7 e^{\mu_8 \phi_2(\mathbf z)} \ge 0 \\ {\mathbf E}_0 (\mathbf z) &= e^{-\mathbf z \cdot \beta_0^D (\mathbf z)}
\end{align}
We update $\{\beta_{t}, {\mathbf E}_t\}$ for $t$ iterations until $\mathcal{L}_z$$\le0.01$.

\begin{figure}[t]
    %\vspace{-2mm}
    \centering
    \begin{subfigure}{0.5\textwidth}
        \includegraphics [width=\linewidth]{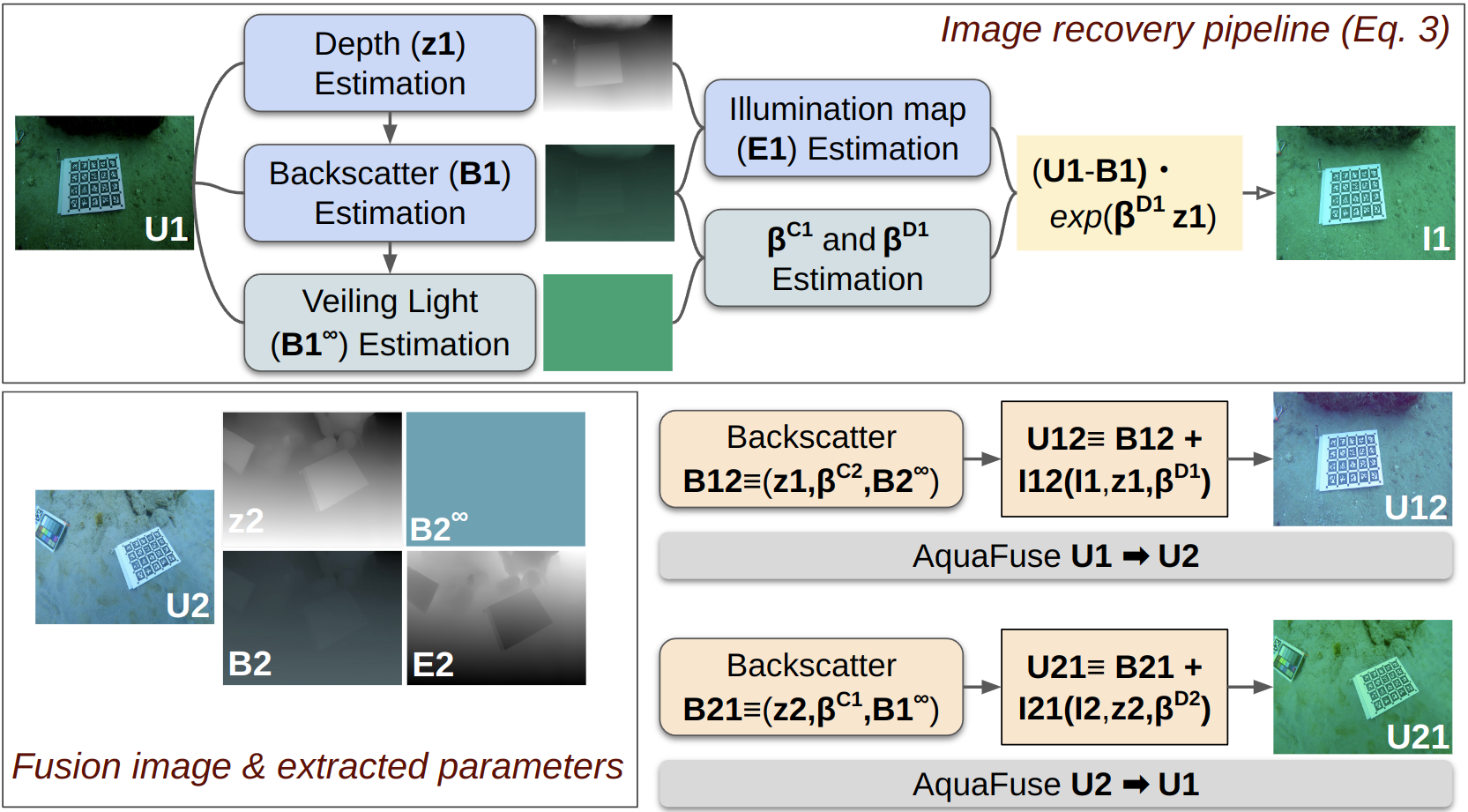}%
        %\vspace{-1mm}
        \caption{Waterbody fusion algorithm of AquaFuse is shown; the input image $\mathbf{U}_1$ is fused with the waterbody from a reference image $\mathbf{U}_2$.}
        \label{fig:fusion}
    \end{subfigure}

    \begin{subfigure}{0.5\textwidth}
    \vspace{2mm}
        \includegraphics [width=\linewidth]{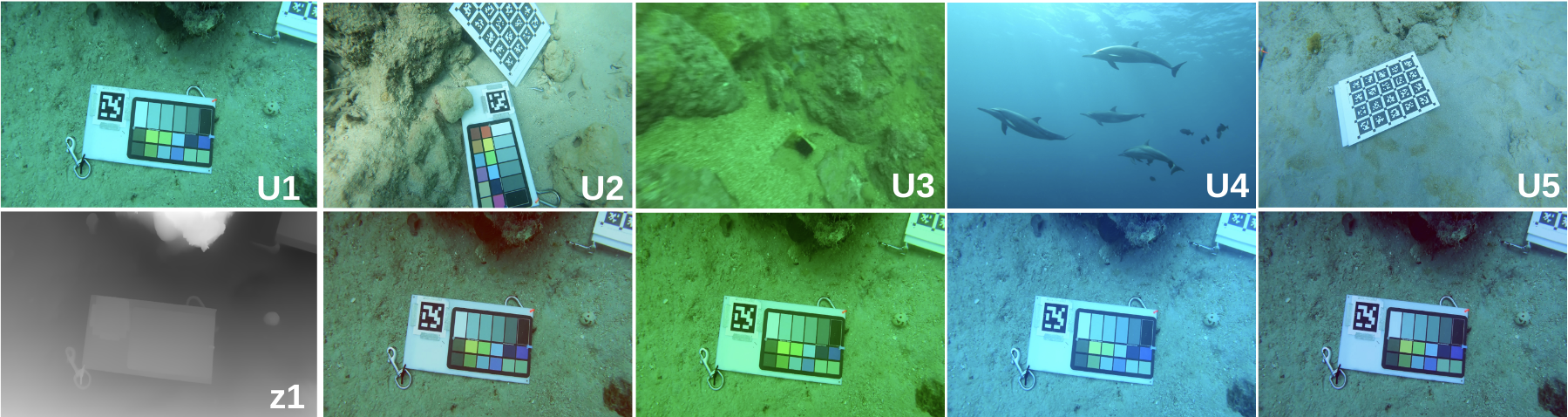}%
        %\vspace{-1mm}
        \caption{More examples are shown where an input image $\mathbf{U}_1$ is fused with the waterbody from reference images $\mathbf{U}_2$ through $\mathbf{U}_5$.}%
        \label{fig:thickness}
    \end{subfigure}%
    \vspace{-1mm}
    \caption{Computational outline of AquaFuse with examples.}
    \label{fig:aquafuse}
    \vspace{-2mm}
\end{figure}

\vspace{1mm}
\noindent
\textbf{Empirical estimation of veiling light}. In practice, \textit{background light} is estimated as a proxy of $\mathbf B_\mathbf \lambda^\infty$ from the intensities of brightest background pixels~\cite{li2016single,li2018emerging}. This assumption is invalid when foreground objects are brighter than the global background light. Various hierarchical searching technique based on quad-tree subdivision or other region-based selection mechanisms~\cite{li2016underwater} are typically used to avoid bright foreground pixels. In our implementation, we exploit the $5\%$ farthest pixels by exploiting the estimated depth map. Specifically, we \textit{mask} $\mathbf{Z}$ to compute the median pixel values of $5\%$ farthest pixels in $\mathbf{U}$. When insufficient ($\le100$) pixels are are found, we perform a region-based selection on $20\%$ pixels on the top-left and top-right corners of the image, followed by bottom corners. Subsequently, we apply a color filter to remove non-background colors within the bluish-green range, ignoring the red channel. This is because the blue light travels the longest distance in the water because of its shortest wavelength, followed by the green light and then the red light.

\subsection{Waterbody Fusion}
Fig.~\ref{fig:fusion} presents the waterbody fusion algorithm of the proposed AquaFuse system. For a input image $\mathbf{U1}$ and reference image $\mathbf{U}2$, we first generate their respective depth maps and waterbody parameters. Then, we compute the \textit{depth factor} of $\mathbf{U1}$ as $\mathbf{F1} = e^{-\beta^{B1}\cdot \mathbf{z1}}$. The subsequent backscatter modulation of the input image is performed as: 
\begin{equation}
    \mathbf{B1}^* = \mathbf{B1} \cdot \mathbf{F1} \cdot \cos\theta
    \label{veil_factor}
\end{equation}
where $\theta$ is the incident angle of the light~\cite{najem2024incidence,seckmeyer1993cosine,wang2021underwater}. Note that $\beta^{B1}$ controls the rate at which water attenuates light, with higher values representing murkier water and leading to faster decay of backscatter~\cite{basri2003lambertian}. Besides, $\theta$ determines how light interacts with the water surface and depth, influencing the amount of backscatter in different regions.

Following the image formation model presented in Eq.~\ref{reconstruct}, we remove the backscatter to get $\mathbf{U1}^*$ as: 
\begin{equation}
    \mathbf{U1}^* = \mathbf{U1} - \mathbf{B1}^* = \mathbf{U1} - \mathbf{B}1\cdot \mathbf{F1} \cdot \cos\theta
\end{equation}
It is important to note that the veiling light is \textit{global}, while the backscatter signal is not; thus, $\mathbf{B2^{\infty}}$ is the only paramter we can directly transfer from $\mathbf{U2}$. We first factorize $\mathbf{B2^{\infty}}$ with the scene depth information of $\mathbf{U1}$, then generate the \textit{fused} backscatter signal $\mathbf{B12}$ by keeping the scene geometry and local structure of $\mathbf{U1}$ intact. Specifically, we perform: 
\begin{align*}
    \mathbf{B2_*^{\infty}} &= \alpha \cdot \mathbf{B2^\infty} \cdot \mathbf{F1} \cdot \cos\theta, \\
    \mathbf{B12} &= \mathbf{B2_*^{\infty}} (1 - e^{-\beta^{B1} \cdot \mathbf{z1} }) .
\end{align*}
Finally, the waterbody fusion is performed by adding the fused backscatter signal to the direct attenuation signal as follows:  
\begin{equation}
    \mathbf{U12} = \mathbf{I1} \cdot e^{-\beta^{D1} \cdot \mathbf{z1}} +  \mathbf{B12}.
    \label{final_eq}
\end{equation}
As shown in Fig.~\ref{fig:fusion}, we can perform an analogous operation of fusing $\mathbf{B1^{\infty}}$ into $\mathbf{B21}$ to generate $\mathbf{U21}$, which will fuse the waterbody of $\mathbf{U1}$ image into the content of $\mathbf{U2}$.

%% file: src/Experiments.tex
\begin{figure}[b]
    %\vspace{-2mm}
    \centering
    \begin{subfigure}{0.5\textwidth}
        \includegraphics [width=\linewidth]{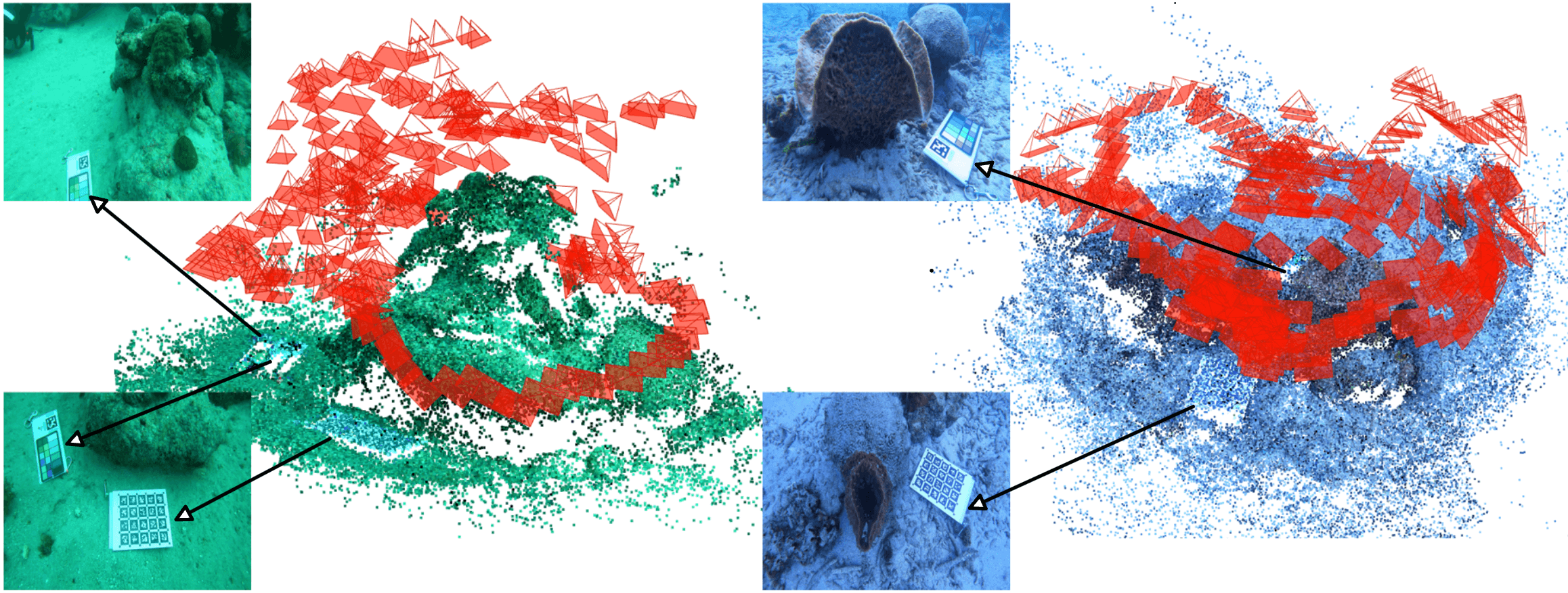}%
        %\vspace{-1mm}
        \caption{3D reconstructions (points) and camera poses (red cones) for two data collection trials at $30'$ and $80'$ depths are shown.}
        \label{fig:3d_data}
    \end{subfigure}

    \begin{subfigure}{0.5\textwidth}
    \vspace{3mm}
        \includegraphics [width=\linewidth]{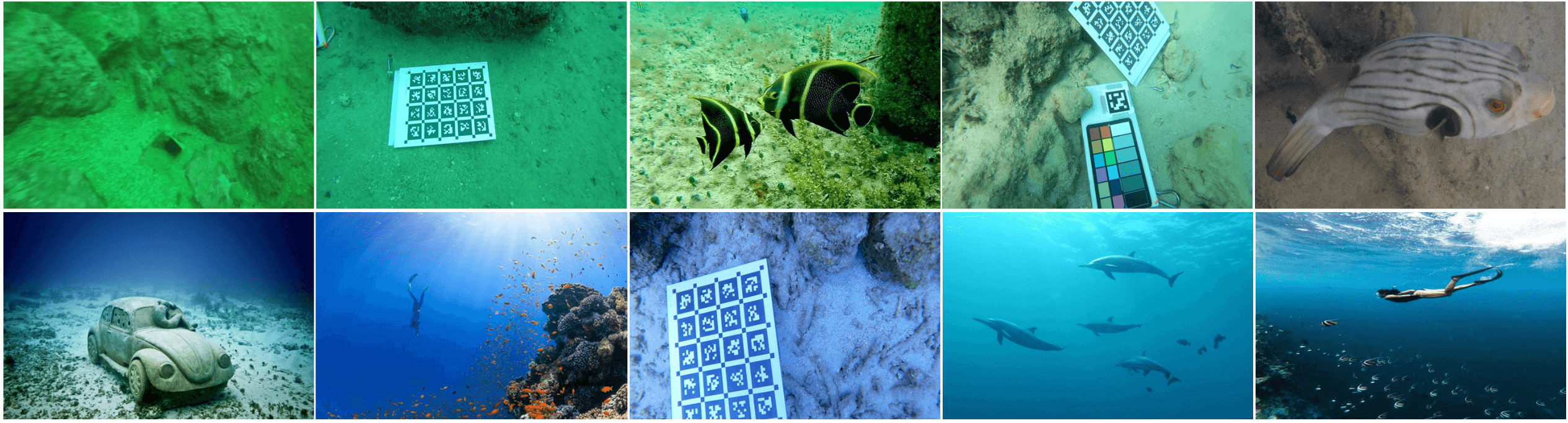}%
        %\vspace{-1mm}
        \caption{Sample reference scenes used for waterbody fusion; these images are in benchmark datasets: USOD~\cite{USOD10K}, UIEB~\cite{li2019underwater}, and EUVP~\cite{islam2020fast}.}%
        \label{fig:ref_samples}
    \end{subfigure}%
    \vspace{-1mm}
    \caption{Our evaluation samples include comprehensive data collected on six underwater sites with incremental depth intervals ($10'$-$80'$). We also use reference images from a few existing benchmark datasets.}
    \label{fig:data_info}
    \vspace{-2mm}
\end{figure}

\section{Performance Evaluation}\label{sec:performance}

\subsection{Setup and Real-world Data}
For experimental evaluation, we collected data at a coral reef site of Bellairs Research Center, Barbados. We used a stereo rig with two GoPros at six different water depths: $10'$, $20'$, $30'$, $50'$, $70'$, and $80'$ approximately. Fig.~\ref{fig:3d_data} shows two sample scenes and corresponding camera poses for our trials. We also used reference images from benchmark underwater datasets such as USOD~\cite{USOD10K}, Sea-thru~\cite{akkaynak2019sea}, UIEB~\cite{li2019underwater}, and EUVP~\cite{islam2020fast}; see Fig.~\ref{fig:ref_samples}. These reference scenes are characterized by varying waterbody conditions provide a diverse range of test cases for performance evaluation.

\begin{table}[t]
    \centering
    \caption{Image similarity scores (PSNR, SSIM) between AquaFuse-generated images averaged across eight different waterbody fusions and raw input, and estimation error (RMSE) of their respective depth-maps are shown for data collected at different depth intervals. Scores are represented as $mean$$\pm$$variance$.}
    \label{table_img}
    \vspace{-1mm}
    \resizebox{\columnwidth}{!}{%
    \begin{tabular}{cccc}
    \toprule
       {Depth} & {PSNR ($\uparrow$)} & {SSIM ($\uparrow$)} & {RMSE ($\downarrow$)}  \\ \midrule
      $10'$  &$27.85\pm0.17$& $0.925\pm0.010$  &  $5.004\pm3.660$  \\
      $20'$ &$27.90\pm0.25$&  $0.952\pm0.003$  &  $4.464\pm2.207$  \\
      $30'$ &$27.86\pm0.10$&  $0.929\pm0.001$  &  $5.250\pm1.995$  \\
      $50'$ &$28.02\pm0.21$&  $0.904\pm0.018$  &  $5.169\pm3.520$  \\
      $70'$ &$28.33\pm0.75$&  $0.924\pm0.017$  &  $5.580\pm4.666$  \\
      $80'$ &$28.36\pm1.04$&  $0.902\pm0.027$  &  $6.603\pm6.437$  \\
      \bottomrule
    \end{tabular}
    }
    \label{tab:my_label}
    \vspace{-2mm}
\end{table}

\begin{figure}[h]
\vspace{-1mm}
    \centering
    \includegraphics [width=\linewidth]{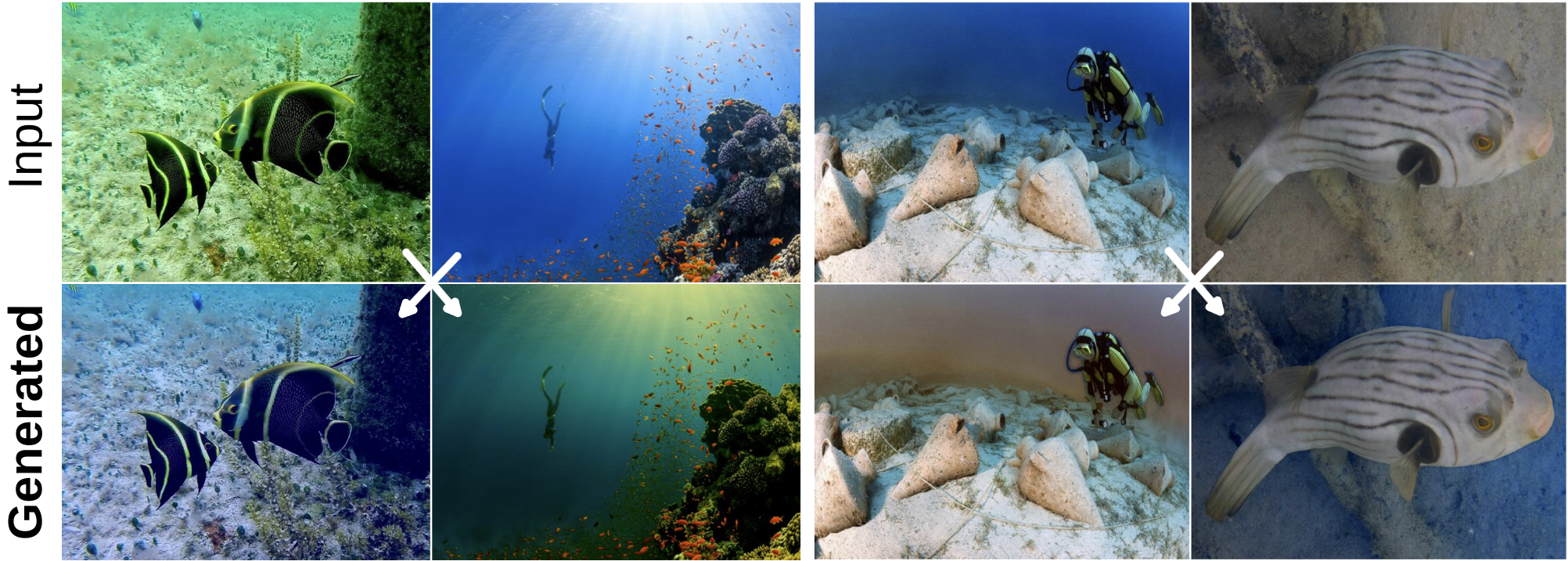}%
    \vspace{-1.5mm}
    \caption{Two examples of \textit{waterbody crossover} by AquaFuse are shown for physics-aware data augmentation; note that this process enables $2\times$ data generation.}%
    \vspace{-4mm}
\label{fig:xover}
\end{figure}

\subsection{Quantitative Evaluation}\label{sec:quant}
Table~\ref{table_img} presents a quantitative evaluation of the AquaFused images compared to corresponding raw input images at varying depth intervals. The PSNR (peak signal-to-noise ratio) and SSIM (structural similarity index measure) scores~\cite{wang2004image} indicate that AquaFuse maintains strong visual fidelity and structural integrity in the fusion process. The PSNR values range between $27.85$ and $28.36$, while SSIM scores remain consistently over $0.90$, \ie, $90$\% of the structural information is preserved in the AquaFused images. These findings are consistent with the depth error scores as well; as shown in Table~\ref{table_img}, the estimated depths from AquaFused images induce about $4$-$6$\% RMSE compared to direct estimation from the input images. The error rates increase by $1$-$5$\% in deeper water from $10'$ to $80'$, which is expected. Nevertheless, over $94$\% depth consistency and $90$-$95$\% structural similarity scores validate the robustness of the waterbody fusion performance of AquaFuse.

\subsection{Qualitative Experiments}
\subsubsection{Waterbody Crossover aka Style Transfer} AquaFuse can be used as a style transfer method for blending \textit{waterbody appearance} between two images. As shown in Fig.~\ref{fig:xover}, image pairs in the top row are used as reference images to each other to \textit{crossover} their waterbodies. Although the objects in these scenes are entirely different (\eg, fish, human diver, coral reef, rocks) and at varied depths, foreground renderings of the fused waterbody are perceptually realistic and natural. These transformations simulate diverse underwater conditions, providing more useful data augmentation results compared to data-driven methods. In particular, traditional domain style transfer methods demonstrate an averaging effect~\cite{li2019underwater,li2018synthesis,yang2022underwater}, and often generate distorted unrealistic images.

\begin{figure*}[t]
    \centering
    \includegraphics [width=\linewidth]{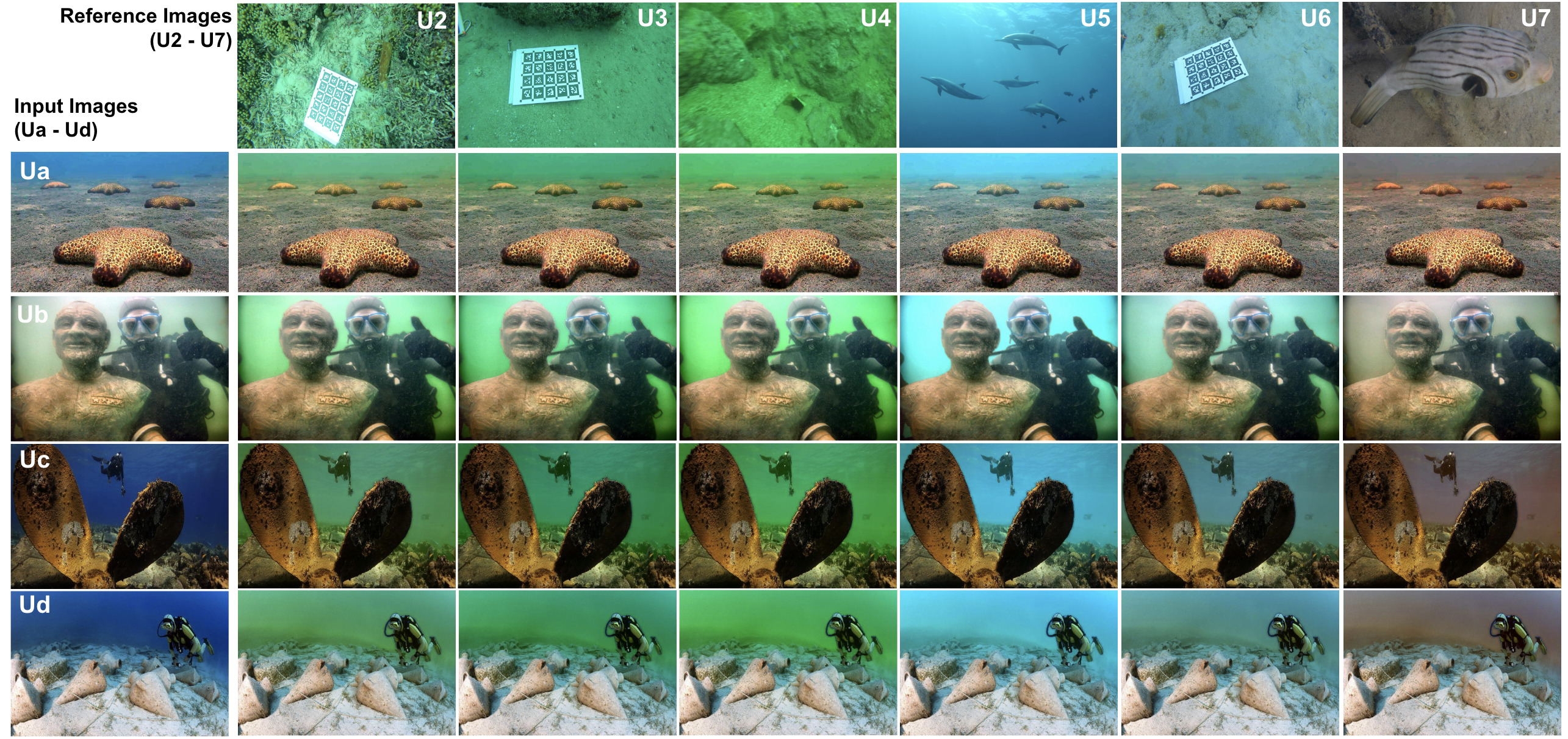}%
    \vspace{-3mm}
     \caption{Perceptually realistic and accurate waterbody fusion capability of AquaFuse are shown for four sample inputs ($\mathbf{Ua}$-$\mathbf{Ud}$) and six diverse reference scenes ($\mathbf{U2}$-$\mathbf{U7}$). Each input image is fused with the varied reference waterbody conditions (style, turbidity, color, and clarity) while preserving its structural integrity of objects (enabling $6\times$ data augmentation).}%
     \vspace{-2mm}
\label{fig:res}
\end{figure*}

\subsubsection{Image Data Augmentation} We now demonstrate how AquaFuse can be used for multi-fold data augmentation of underwater image databases. The proposed waterbody fusion is a unique form of photometric transformation, beyond traditional perspective or isometric functions applied for altering brightness, color, contrast, etc. We validate this by using a few reference images with diverse hues, colors, and contrast variations. We then apply AquaFuse with these reference scenes to sample images from benchmark datasets~\cite{USOD10K,li2019underwater,islam2020fast} for physics-aware waterbody fusion. 

A few examples of $6\times$ data augmentation by AquaFuse are shown in Fig.~\ref{fig:res}; we select six reference images ($\mathbf{U2}$-$\mathbf{U7}$), which are fused to input samples ($\mathbf{Ua}$-$\mathbf{Ud}$). As shown in each row, input images are fused to six reference images' waterbodies while preserving their original scene contents. The columns are arranged to show waterbody fusion to greenish ($\mathbf{U2}$,$\mathbf{U3}$,$\mathbf{U4}$), blueish ($\mathbf{U5}$,$\mathbf{U6}$), and clear water ($\mathbf{U7}$). A total of $6\times4$=$24$ AquaFused images are generated, demonstrating its utility and effectiveness.

In particular, we find that the close-up objects (in $\mathbf{Ub}$,$\mathbf{Uc}$) remain sharp, undistorted, and accurately blended with the surrounding waterbody. The dark shadow regions (in $\mathbf{Uc}$) or bright foreground objects (in $\mathbf{Ud}$) remain intact through the fusion process. Overall, AquaFuse effectively fuses the waterbody into a given image, generating diverse scenarios while preserving the fine object-level details in the scene. One can use more variety of reference images for comprehensive data augmentation of underwater image databases; since the object scene geometry remains preserved, standard annotations (for object detection, segmentation, etc.) would remain the same, thus facilitating a multi-fold increase in training samples.

\subsubsection{Image Enhancement} 
\begin{figure}[h]
%\vspace{-2mm}
    \centering
    \includegraphics [width=\linewidth]{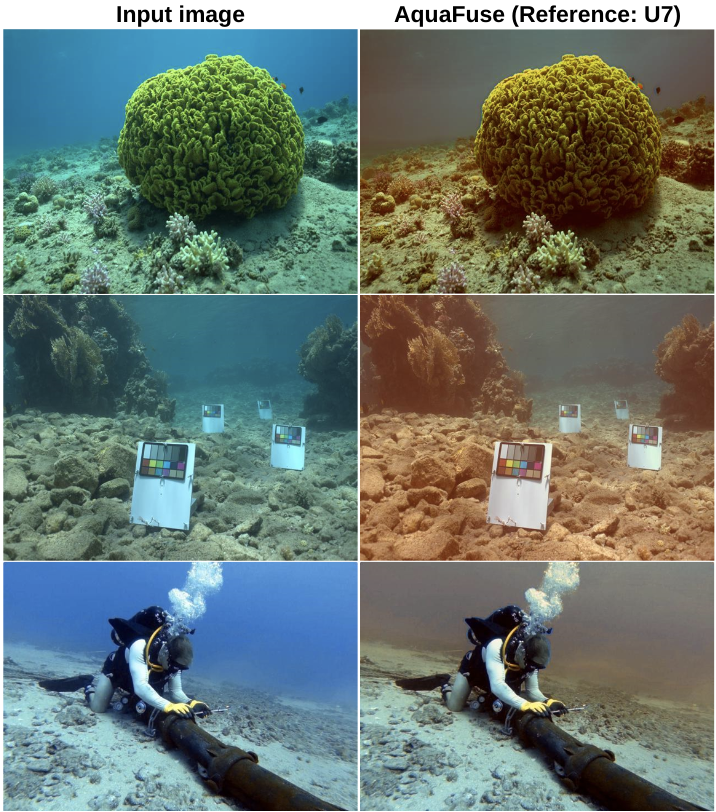}%
    \vspace{-1.5mm}
    \caption{A positive consequence of using AquaFuse with high-quality clear reference scene is that it renders perceptually enhanced images. As shown, the AquaFused images are almost de-watered, and have significantly better image statistics in terms of color, contrast, and sharpness.}%
    \vspace{1mm}
\label{fig:enhance}
\end{figure}

Another useful feature of AquaFuse is that it can transfer \textit{clear water} styles as well, which essentially results in clear fused images. As shown in the last column in Fig.~\ref{fig:res}, with a clear high-resolution reference image ($\mathbf{U7}$), all input images $\mathbf{Ua}$, $\mathbf{Ub}$, $\mathbf{Uc}$, and $\mathbf{Ud}$ -- generated almost dewatered enhanced images. The \textit{transparent} nature of the reference image is fused accurately, resulting in crisp and tint-free images~\cite{islam2020sesr,islam2020fast}. This fusion reduces the impact of the hazy and scattered \textit{veil} in all input images, which validates our intuition of factorizing veiling light to drive the waterbody fusion process (see Eq.~\ref{veil_factor}-\ref{final_eq}). This feature has significant use cases for applications requiring high-quality water-free images, especially in mapping coral reefs and other marine biological samples~\cite{modasshir2021autonomous,alonso2019coralseg,manderson2018vision} and 3D reconstruction~\cite{wu2023sdu_sfm,yang2024seasplat}. A few more examples are illustrated in Fig.~\ref{fig:enhance}. 

\begin{figure}[h]
\vspace{-2mm}
    \centering
    \includegraphics [width=0.98\linewidth]{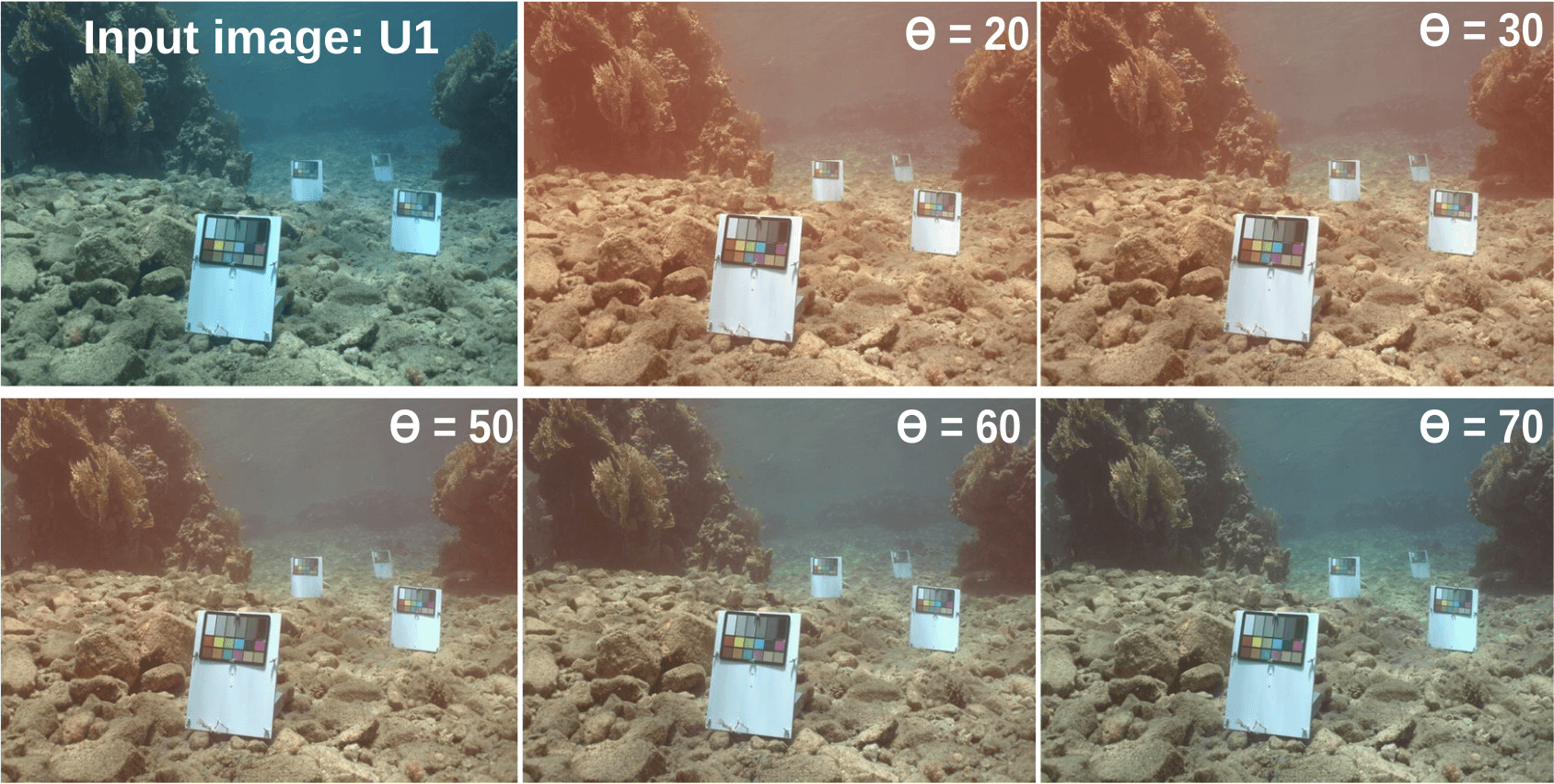}%
    \vspace{-1.5mm}
    \caption{Effects of incident angle $\theta$ on AquaFuse are shown (with reference $\mathbf U7$); lower values of $\theta$ cause more reddish tones (from reference image), while higher values reduce the fusion effects.
    }%
    %\vspace{-1mm}
\label{fig:incident}
\end{figure}

In Fig.~\ref{fig:incident}, we further demonstrate the influence of $\theta$, the incident angle between incoming light rays and the surface normal. While an exact estimation of $\theta$ is challenging outside controlled environments, this relationship is generally represented using a scalar cosine factor $\cos\theta$ corresponding to the \textit{light incidence}~\cite{najem2024incidence,seckmeyer1993cosine,wang2021underwater} amount. As Fig.~\ref{fig:incident} shows, the interpretation for AquaFuse is that $\theta$ can be tuned to control from $\theta\rightarrow0$ (maximum fusion) to $\theta\rightarrow90^\circ$ (minimum fusion) of waterbody styles from reference image to the content image. Our empirical tests show that $\theta \in [30^\circ,60^\circ]$ generates consistent fusion for most natural underwater scenes. 

\begin{figure*}[t]
    \centering
    \includegraphics [width=\linewidth]{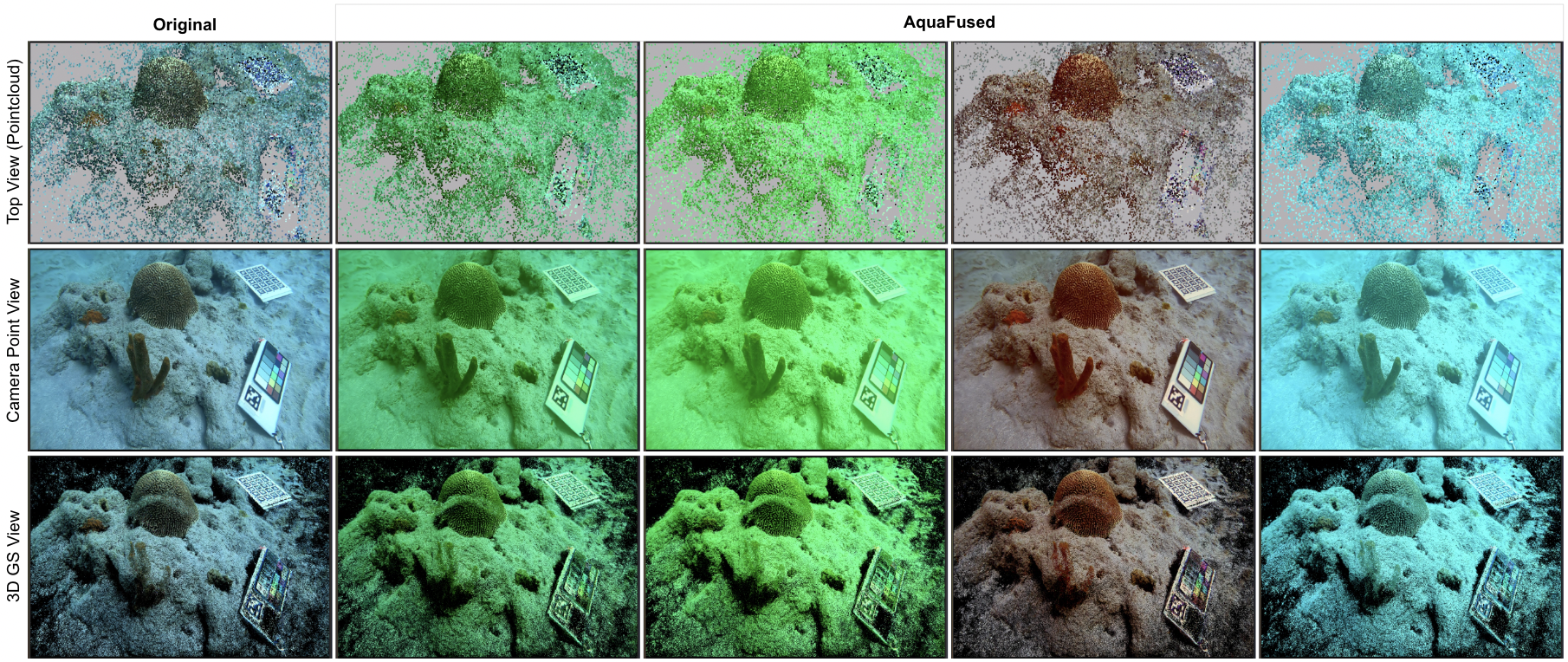}%
    \vspace{-2mm}
    \caption{Geometric validity of AquaFuse is shown with 3D rendering by Gaussian Splatting for four sets of AquaFused scenes; the object structure and scene geometry remain intact as the original input scene, while the waterbody characteristics are fused accurately. See the full video demonstration here: \url{https://youtu.be/gj3QFDMY9Ck}.
    }%
\label{fig:res3d}
\end{figure*}

\subsubsection{Geometry-preserving Scene Transformation}
We further validate that the proposed waterbody fusion process preserves the geometric integrity of the generated scenes. To assess this, we compare the 3D reconstructed scenes from AquaFused images against those generated from the original input images. The rendered views are generated using the Gaussian Splatting (GS) algorithm~\cite{kerbl20233d}, representing scenes as a set of \textit{splats} -- which can be visualized at different viewpoints, scales, and densities. For visualization, we use the SIBR viewer~\cite{philip2019multi,chaurasia2013depth} (version $0.9.6$) to generate a 3D scene rendering. These views allow us to compare the 3D geometric consistency of AquaFused images with the original scenes.

As shown in Fig.~\ref{fig:res3d}, AquaFuse effectively preserves geometric integrity while performing waterbody fusion. The figure compares \textit{3D top view}, \textit{camera point view}, and \textit{GS view} of scenes reconstructed from original and AquaFused images. These point clouds and scene representations reveal that the objects (\eg, coral reefs, seabed, and calibration chart) retain their shape and detail. AquaFuse transfers the waterbody properties without altering the scene geometry, maintaining the underwater environments' realism in the waterbody fusion process. A more descriptive demonstration is in \url{https://youtu.be/gj3QFDMY9Ck}.

These visual inspections corroborate our quantitative analyses of the reconstructed scenes. As shown in Table~\ref{table_aqua3d}, the PSNR, SSIM, and Learned Perceptual Image Patch Similarity (LPIPS) scores of AquaFused images are within $1$-$2$\% of original scenes. In fact, AquaFuse slightly improves 3D reconstruction quality with a higher PSNR and lower LPIPS at shallow depths ($10'$-$30'$). For deep-water scenes (over $60'$), the scores are very similar to using the original images' reconstruction quality, which validates our intuition and design objectives. Additionally, 3D reconstructions with AquaFused images show minimal geometric deviations, measured based on the Surface Normal Deviation (SND) metric. As shown in Table~\ref{table_aqua3d}, the SND is less than $0.3\%$ compared to the original scenes, validating the geometric accuracy of AquaFuse.

\begin{table}[h]
\vspace{1mm}
    \centering
    \caption{3D reconstruction (Gaussian Splatting) performance of AquaFused images versus original images at varied water depths, based on PSNR ($\uparrow$), SSIM ($\uparrow$), and SND ($\downarrow$).
    }
    \label{table_aqua3d}
    %\footnotesize
    \vspace{-1mm}
    \resizebox{0.95\columnwidth}{!}{%
    \begin{tabular}{llccccr}
    \toprule
       {\textbf{Depth}} & {\textbf{Image}} & {\textbf{PSNR}} & {\textbf{SSIM}} & {\textbf{LPIPS}} & {\textbf{SND} ($\%$)} \\ 
       \midrule
      \multirow{2}{*}{$10'$}  & Original   & $22.83$ & $0.8104$  &  $0.2289$  & \multirow{2}{*}{$0.29$} \\
             & AquaFused  & $24.32$ & $0.8137$  &  $0.2073$ &  &  \\ 
             \hline
      \multirow{2}{*}{$20'$}  & Original   & $25.32$ & $0.8154$  &  $0.2282$ & \multirow{2}{*}{$0.002$} \\
             & AquaFused  & $27.13$ & $0.8031$  &  $0.2232$ &  &  \\ 
             \hline
      \multirow{2}{*}{$30'$}  & Original   & $25.03$ & $0.8041$  &  $0.2060$  & \multirow{2}{*}{$0.25$} \\
             & AquaFused  & $25.39$ & $0.7872$  &  $0.2044$ &  &  \\ 
             \hline
      \multirow{2}{*}{$50'$}  & Original   & $23.82$ & $0.7587$  &  $0.2013$ & \multirow{2}{*}{$0.03$} \\
             & AquaFused  & $24.74$ & $0.7619$  &  $0.1984$ &  &  \\ 
             \hline
      \multirow{2}{*}{$70'$}  & Original   & $24.13$ & $0.7431$& $0.2768$ & \multirow{2}{*}{$0.18$} \\
             & AquaFused  & $23.78$&$0.7360$& $0.2950$ &  &  \\ 
             \hline
      \multirow{2}{*}{$80'$}  & Original   & $23.91$ & $0.7320$  &  $0.3225$ & \multirow{2}{*}{$0.29$} \\
             & AquaFused  & $23.01$ & $0.7231$  &  $0.3498$ &  &  \\ 
      \bottomrule
    \end{tabular}
    }
    %\vspace{-3mm}
\end{table}

\vspace{1mm}
\noindent
\textbf{Computational performance analysis.} AquaFuse offers a closed-form solution requiring only standard OpenCV libraries, thus offering fast end-to-end inference. As Table~\ref{table_fps_memory} indicates, AquaFuse offers $342.3$\,FPS ($20.69$\,FPS) inference rates for $256\times256$ ($1920\times1080$) image resolutions on a basic CPU -- which is several orders of magnitude higher than SOTA models. Even on a resource-constrained Raspberry Pi-4 device, AquaFuse offers $22.58$ FPS inference ($44.3$\,milliseconds/image) -- which no data-driven style transfer methods can offer. These results validate AquaFuse's feasibility for real-time use on marine robots and standalone sensor nodes.

\begin{table*}[t]
    \centering
    \caption{An extension of Table~\ref{tab:my_label}, for quantitatively comparing AquaFuse with SOTA style transfer methods. The SSIM and RMSE scores are provided as $mean$$\pm$$variance$.
    }
    \label{table_img_models}
    \vspace{-1mm}
    \resizebox{1.5\columnwidth}{!}{ % Resize the table to fit within the column width
    \begin{tabular}{ccccccc}
    \toprule
       {Depth} & \multicolumn{2}{c}{{ASTMAN~\cite{deng2020arbitrary}}} & \multicolumn{2}{c} { {IEContraAST~\cite{chen2021artistic}}} & \multicolumn{2}{c}{{MCCNet~\cite{deng2021arbitrary}}} \\ 
       \cmidrule(lr){2-3} \cmidrule(lr){4-5} \cmidrule(lr){6-7}
       & {SSIM ($\uparrow$)} & {RMSE ($\downarrow$)} 
       & {SSIM ($\uparrow$)} & {RMSE ($\downarrow$)} 
       & {SSIM ($\uparrow$)} & {RMSE ($\downarrow$)} \\ \midrule
      $10'$  & $0.28\pm0.02$ & $51.61\pm4.69$ & $0.59\pm0.06$ & $32.11\pm5.09$ & $0.54\pm0.06$ & $28.63\pm5.74$ \\
      $20'$  & $0.35\pm0.02$ & $33.02\pm0.72$ & $0.64\pm0.04$ & $24.61\pm1.29$ & $0.63\pm0.02$ & $22.05\pm1.71$ \\
      $30'$  & $0.21\pm0.02$ & $54.08\pm2.22$ & $0.62\pm0.03$ & $23.09\pm2.17$ & $0.57\pm0.01$ & $20.81\pm1.99$ \\
      $50'$  & $0.16 \pm 0.01$ & $51.23\pm3.77$ & $0.59\pm0.05$ & $25.86\pm4.29$ & $0.40\pm0.12$ & $29.21\pm3.87$ \\
      $70'$  & $0.27 \pm 0.03$ & $59.45 \pm 6.46$ & $0.69 \pm 0.05$ & $29.48 \pm 4.38$ & $0.58 \pm 0.14$ & $31.23 \pm 4.32$ \\
      $80'$  & $0.36\pm0.02$ & $58.82\pm7.07$ & $0.75\pm0.05$ & $34.71\pm4.66$ & $0.65\pm0.08$ & $32.21\pm5.03$ \\
      \midrule
       {Depth} & \multicolumn{2}{c}{{StyTr2~\cite{deng2022stytr2}}} & \multicolumn{2}{c}{{PhotoWCT2~\cite{chiu2022photowct2}}} & \multicolumn{2}{c}{\textbf{{AquaFuse}}} \\ 
       \cmidrule(lr){2-3} \cmidrule(lr){4-5} \cmidrule(lr){6-7}
       & {SSIM ($\uparrow$)} & {RMSE ($\downarrow$)} 
       & {SSIM ($\uparrow$)} & {RMSE ($\downarrow$)} 
       & {SSIM ($\uparrow$)} & {RMSE ($\downarrow$)} \\ \midrule
      $10'$  & $0.30\pm0.02$ & $43.29\pm1.93$ & $0.82\pm0.07$ & $16.16\pm1.48$ & $0.93\pm0.01$ & $5.00\pm3.66$ \\
      $20'$  & $0.36\pm0.03$ & $27.37\pm0.44$ & $0.88\pm0.04$ & $15.74\pm1.58$ & $0.95\pm0.00$ & $4.46\pm2.21$ \\
      $30'$  & $0.22\pm0.03$ & $42.16\pm0.39$ & $0.79\pm0.05$ & $15.26\pm1.57$ & $0.93\pm0.00$ & $5.25\pm1.99$ \\
      $50'$  & $0.16\pm0.02$ & $41.91\pm1.55$ & $0.76\pm0.06$ & $14.67\pm2.19$ & $0.90\pm0.02$ & $5.17\pm3.52$ \\
      $70'$  & $0.31 \pm 0.03$ & $44.17 \pm 3.33$ & $0.82 \pm 0.06$ & $17.79 \pm 2.44$ & $0.92\pm0.02$ & $5.58\pm4.67$ \\
      $80'$  & $0.38\pm0.02$ & $ 45.83\pm2.65$ & $0.87\pm0.03$ & $22.12\pm1.33$ & $0.90\pm0.03$ & $6.60\pm6.44$ \\
      \bottomrule
    \end{tabular}
    }
    %\vspace{-2mm}
\end{table*}

\begin{figure*}[t]
    \centering
    \includegraphics [width=0.96\linewidth]{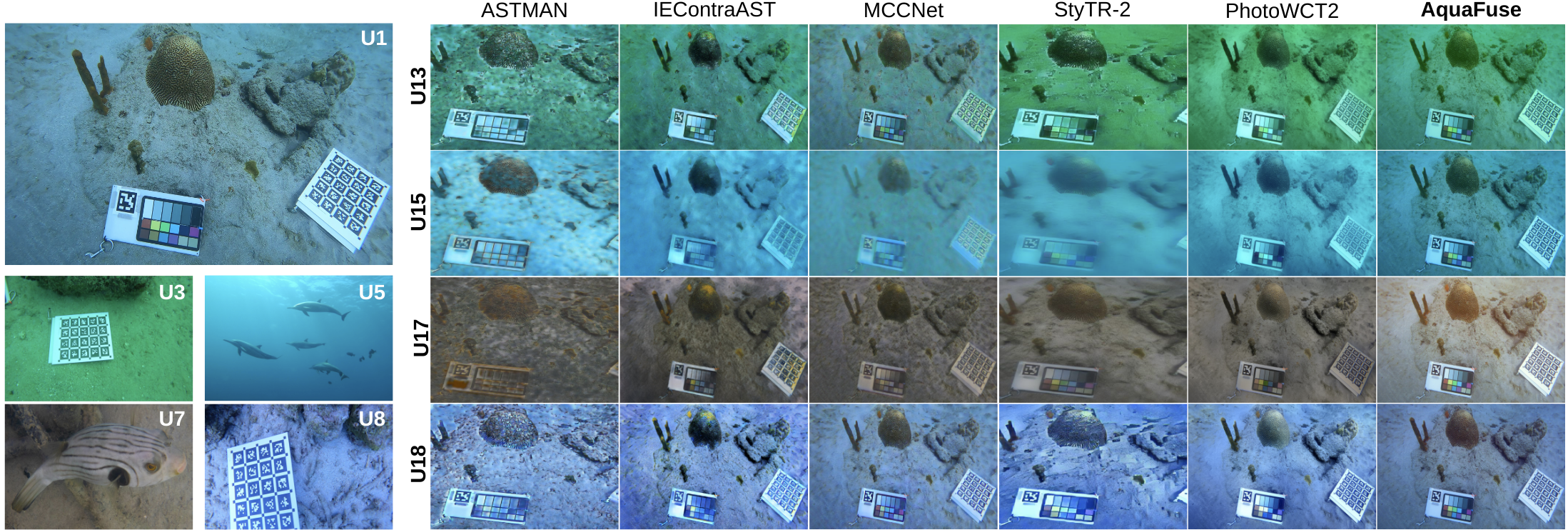}%
    \vspace{-1mm}
     \caption{Comparative qualitative analysis of AquaFuse with SOTA style transfer methods. The input image (content: $\mathbf U1$) is transformed by style images ($\mathbf U3$, $\mathbf U5$, $\mathbf U7$, $\mathbf U8$) at each row.  AquaFuse effectively transfers waterbody style while preserving scene structure.}
\label{fig:comparative}
\vspace{-2mm}
\end{figure*}

\subsection{Comparative Analysis with Traditional Learning-based Style Transfer Methods}
We further investigate SOTA data-driven models for learning waterbody style transfer for three classes (greenish, blueish, and clear) shown in Fig.~\ref{fig:res}. Results reported in Fig.~\ref{fig:comparative} reveal that transformers and CNN-based models (StyTr2~\cite{deng2022stytr2}, MCCNet~\cite{deng2021arbitrary},  ASTMAN~\cite{deng2020arbitrary}) struggle to preserve the structural integrity of the scene. Although the geometry-aware PhotoWCT2~\cite{chiu2022photowct2} demonstrates better performance, its generated images exhibit saturated hues and pixel distortions. 
All models were pre-trained on $118$\,K MSCOCO-2017 samples, then fine-tuned on an additional $1,742$ underwater images per class based on their recommended training settings. 
Quantitative analyses performed on depth-wise datasets corroborate these observations; see Table~\ref{table_img_models}. AquaFuse significantly outperforms competing models across all metrics, validating the robustness and accuracy of its physics-based formulation.

\begin{table}[h]
    \centering
    \caption{Inference rates in FPS ($\uparrow$) for six SOTA methods across various image resolutions on an Intel\textsuperscript{\textregistered} Core\texttrademark{}-i7 CPU with $16$\,GB RAM; `$\times$' represents out-of-memory (or below $0.01$ FPS) cases.
    }
    \label{table_fps_memory}
    \vspace{-1mm}
    \resizebox{\columnwidth}{!}{ % Resize the table to fit within the column width
    \begin{tabular}{l@{\hskip 2mm}r@{\hskip 2mm}r@{\hskip 2mm}r@{\hskip 2mm}r}

    \toprule
       {Resolution} & {$256\times256$} & {$512\times512$} & {$1024\times1024$} & {$1920\times1080$} \\ 
       \midrule
      ASTMAN~\cite{deng2020arbitrary}        & $0.52$    & $0.10$    & $\times$    & $\times$ \\
      IEContraAST~\cite{chen2021artistic}    & $0.79$    & $0.17$    & $0.03$    & $\times$ \\
      MCCNet~\cite{deng2021arbitrary}        & $0.48$    & $0.11$    & $0.03$    & $\times$ \\
      StyTr2~\cite{deng2022stytr2}       & $0.12$    & $\times$    & $\times$   & $\times$ \\
      PhotoWCT2~\cite{chiu2022photowct2}     & $0.75$    & $0.21$    & $0.04$    & $0.03$ \\
      Sea-thru~\cite{akkaynak2019sea}      & $1.22$    & $0.41$    & $0.11$    & $0.05$ \\
      \textbf{AquaFuse} & $342.30$  & $84.37$   & $20.69$   & $9.53$ \\
      \bottomrule
    \end{tabular}
    }
    %\vspace{-4mm}
\end{table}

\subsection{Challenges \& Practicality}
Despite the robust performance, there are some practicalities and challenges involved in the waterbody fusion process of AquaFuse.  First, accurate background light estimation for scenes with variable depth levels is crucial to maintaining fusion quality. Our region-based selection often fails when all pixels in the image have very similar depths; this is a common issue of DCP~\cite{he2010single}-inspired filters for backscatter or background light estimation methods. Moreover, when the attenuation coefficients, incident angle, and other camera parameters are unknown, empirical estimations of backscatter and illumination maps often fail, especially in noisy and turbid scenes. We use an adaptive illumination map estimator guided by depth factor and incident angle to adjust to dynamically varying light conditions, enabling AquaFuse to preserve object geometry even in low visibility. Finally, we balanced the noise and sharpness using a bilateral filter to ensure that essential scene details were maintained. While these special cases can be addressed by manual post-processing, AquaFuse generates robust, accurate, and perceptually realistic waterbody synthesis for natural underwater scenes.

%% file: src/Conclusion.tex
\vspace{-1mm}
\section{Conclusion \& Future Work} \label{sec:conclusion}
\vspace{-1mm} 
This paper introduces AquaFuse, a physics-guided waterbody fusion method for accurate data augmentation, enhancement, and view synthesis of underwater imagery. AquaFuse generates realistic and geometrically consistent underwater images across various depths and waterbody conditions by exploiting the physical characteristics of light scattering, absorption, and backscatter. This makes it a powerful tool for augmenting training databases for underwater image recognition and view synthesis. Another key strength of AquaFuse is that it can handle complex waterbody properties such as varying color, turbidity, and clarity without compromising depth consistency or object geometry. Our experimental results demonstrate that AquaFuse can preserve over 94\% depth consistency and 90-95\% structural similarity in its waterbody fusion process. Moreover, AquaFuse generates accurate 3D reconstructions using techniques like Gaussian Splatting, making it an ideal tool for improving underwater scene diversity and realism. Our future efforts will focus on deploying AquaFuse in the autonomy pipeline of underwater robots for vision-based mapping and active servoing applications.

\vspace{1mm}
\section*{Acknowledgments} 
\vspace{-1mm}
This work is supported in part by the NSF grants \#$2330416$ and \#$1943205$; and UF research grant \#$132763$.